\documentclass[11pt]{article}
\usepackage{naaclhlt2013}
\usepackage{times}
\usepackage{latexsym}
\usepackage{amsmath}
\usepackage{multirow}
\usepackage{url}
\usepackage{graphics}

\usepackage{enumerate}

\title{NRC-Canada: Building the State-of-the-Art in\\ Sentiment Analysis of Tweets}

 \author{Saif M. Mohammad, Svetlana Kiritchenko, and Xiaodan Zhu\\
  	National Research Council Canada\\
 	 Ottawa, Ontario, Canada K1A 0R6\\
   {\tt \{saif.mohammad,svetlana.kiritchenko,xiaodan.zhu\}@nrc-cnrc.gc.ca} }

\date{}

\begin{document}
\maketitle
\begin{abstract}
In this paper, we describe how we created two state-of-the-art SVM classifiers,
one to detect the sentiment of messages such as tweets and SMS (message-level task) and one to detect the
sentiment of a term within a message (term-level task). 
Among submissions from 44 teams in a competition, our submissions stood first in both tasks on tweets,
obtaining an F-score of 69.02 in the message-level task and 88.93 in
the term-level task. We implemented a variety of surface-form, semantic, and
sentiment features. We also generated two large word--sentiment association
lexicons, one from tweets with sentiment-word hashtags, and one from tweets with
emoticons. 
In the message-level task, the lexicon-based  features provided a gain of 5 F-score points over all others.
Both of our systems can be
replicated using freely available resources.\footnote{The three authors contributed equally to this paper. Svetlana
Kiritchenko developed the system for the message-level task, Xiaodan Zhu
developed the system for the term-level task, and Saif Mohammad led the overall effort, co-ordinated both
tasks, and contributed to feature development.}
\end{abstract}

\section{Introduction}
Hundreds of millions of people around the world actively use microblogging websites such as Twitter.
Thus there is tremendous interest in sentiment analysis of tweets 
across a variety of domains such as commerce \cite{Jansen09}, health \cite{Chew10,Salathe11}, and disaster management \cite{Verma11,Mandel12}.


In this paper, we describe how we created two state-of-the-art SVM classifiers,
one to detect the sentiment of messages such as tweets and SMS (message-level task) and one to detect the
sentiment of a term within a message (term-level task). 
The sentiment can be one out of three possibilities: positive, negative, or neutral.
We developed these
classifiers to participate in an international competition organized by the Conference on
Semantic Evaluation Exercises (SemEval-2013) \cite{SemEval2013task2}.\footnote{http://www.cs.york.ac.uk/semeval-2013/task2}
The organizers created and shared sentiment-labeled tweets for training, development, and testing.
The distributions of the labels in the different datasets is shown in Table \ref{tab:dist}.
The competition, officially referred to as {\it Task 2: Sentiment Analysis in Twitter},
had 44 teams (34 for the message-level task and 23 for the term-level task). Our submissions stood first in both tasks,
obtaining a macro-averaged F-score of 69.02 in the message-level task and 88.93 in
the term-level task. 

The task organizers also provided a second test dataset, composed of Short Message Service (SMS) messages
(no training data of SMS messages was provided).
We applied our classifiers on the SMS test set without any further tuning.
Nonetheless, the classifiers still obtained the first position
in identifying sentiment of SMS messages (F-score of 68.46) and second position
in detecting the sentiment of terms within SMS messages (F-score of 88.00, only
0.39 points behind the first ranked system).

We implemented a number of surface-form, semantic, and
sentiment features. We also generated two large word--sentiment association
lexicons, one from tweets with sentiment-word hashtags, and one from tweets with
emoticons. The automatically generated lexicons were particularly useful. In the
message-level task for tweets, they alone provided a gain of more than 5 F-score points over
and above that obtained using all other features. 
The lexicons are made freely available.\footnote{www.purl.com/net/sentimentoftweets}

\begin{table}[t]
\caption{Class distributions in the training set (Train), development set (Dev) and testing set (Test).
The Train set was accessed through tweet ids and a download script. However,
not all tweets were accessible. Below is the number of Train examples
we were able to download. The Dev and Test sets were provided by FTP.} 
\label{tab:dist}
\begin{center}
\resizebox{0.49\textwidth}{!}{
\vspace*{-2mm}
\begin{tabular}{lrrrr}
\hline
	{\bf Dataset}	&{\bf Positive}	&{\bf Negative}	&{\bf Neutral} &{\bf Total}\\
\hline
	{\bf Tweets} & & & \\
	\multicolumn{3}{l}{Message-level task:} &\\
	$\;\;\;$ Train &3,045 (37\%)       &1,209 (15\%)      &4,004 (48\%)    &8,258\\
	$\;\;\;$ Dev & 575	(35\%)& 340 (20\%)	& 739 (45\%)	& 1,654\\
	$\;\;\;$ Test &	1,572 (41\%)	& 601 (16\%)	& 1,640 (43\%) &	3,813\\
	\multicolumn{3}{l}{Term-level task:} &\\
	$\;\;\;$ Train &4,831 (62\%)  &2,540 (33\%) &385 (5\%) &7,756\\
	$\;\;\;$ Dev &648 (57\%) 	&430 (38\%) &57 (5\%)	&1,135\\
	$\;\;\;$ Test &2,734 (62\%)   &1,541 (35\%)    &160 (3\%)	&4,435\\ [5pt]
	{\bf SMS}	& & &\\
	\multicolumn{3}{l}{Message-level task:} &\\
	$\;\;\;$ Test &	492 (23\%) &	394 (19\%)	& 1,208 (58\%) &	2,094\\
	\multicolumn{3}{l}{Term-level task:} &\\
	$\;\;\;$ Test &1,071 (46\%) &1,104 (47\%) &159 (7\%)	&2,334\\
\hline
\end{tabular}
}
\end{center}
\vspace*{-6mm}
\end{table}


\section{Sentiment Lexicons}
Sentiment lexicons are lists of words with associations to positive and negative sentiments.

\subsection{Existing, Manually Created Lexicons}
The manually created lexicons we used include the NRC Emotion Lexicon \cite{MohammadT10,MohammadY11} (about 14,000 words), the MPQA Lexicon \cite{Wilson05} (about 8,000 words), 
and the Bing Liu Lexicon \cite{Hu04} (about 6,800 words).


\subsection{New, Tweet-Specific, Automatically Generated Sentiment Lexicons}

\subsubsection{NRC Hashtag Sentiment Lexicon}

Certain words in tweets are specially marked with a hashtag (\#) to indicate the topic
or sentiment.  Mohammad \shortcite{Mohammad12} showed that hashtagged emotion words such as joy, sadness, angry, and surprised
are good indicators that the tweet as a whole (even without the hashtagged emotion word) is expressing the same emotion.
We adapted that idea to create a large corpus of positive and negative tweets.

We polled the Twitter API every four hours from April to December 2012 in search of tweets with either a positive word hashtag
or a negative word hashtag. 
A collection of 78 seed words closely related to {\it positive} and {\it negative} such as {\it \#good, \#excellent, \#bad,} and {\it \#terrible} were used (32 positive and 36 negative). 
These terms were chosen from entries for {\it positive} and {\it negative} in the Roget's Thesaurus.

A set of 775,000 tweets were used to generate a large word--sentiment association lexicon. 
A tweet was considered positive if it had one of the 32 positive hashtagged seed words, and negative if it had one of the 36
negative hashtagged seed words.
The association score for a term $w$ was calculated from these pseudo-labeled tweets as shown below:
{\small
\begin{equation}
score (w) = PMI(w,positive) - PMI (w, negative)
\end{equation}
}
\noindent where PMI stands for pointwise mutual information. A positive score indicates
association with positive sentiment, whereas a negative score indicates association with negative sentiment.
The magnitude is indicative of the degree of association.
The final lexicon, which we will refer to as the {\it NRC Hashtag Sentiment Lexicon} has entries for 54,129 unigrams
and  316,531 bigrams.
Entries were also generated for unigram--unigram, unigram--bigram, and bigram--bigram pairs that were 
not necessarily contiguous in the tweets corpus. Pairs with certain punctuations, `@' symbols, and some function words were removed.
The lexicon has entries for 308,808 non-contiguous pairs.

 \subsubsection{Sentiment140 Lexicon}

The sentiment140 corpus \cite{Go2009} is a collection of 1.6 million tweets that contain positive and negative emoticons. 
The tweets are labeled positive or negative according to the emoticon. 
We generated a sentiment lexicon from this corpus
in the same manner as described above (Section 2.2.1).
This lexicon has entries for 62,468 unigrams, 677,698 bigrams, and 480,010 non-contiguous pairs.

\section{Task: Automatically Detecting the Sentiment of a Message}
The objective of this task is to determine whether a given message is positive, negative, or neutral.

\subsection{Classifier and features}
We trained a Support Vector Machine (SVM) \cite{liblinear} on the training data provided.
SVM is a state-of-the-art learning algorithm proved to be effective on text categorization tasks and robust on large feature spaces. 
The linear kernel and the value for the parameter C=0.005 were chosen by cross-validation on the training data.

We normalized URLs to http://someurl and userids to @someuser. We tokenized and part-of-speech tagged the tweets with the Carnegie Mellon University (CMU) tool \cite{Gimpel11}.
Each tweet was represented as a feature vector made up of the following groups of features:
\begin{itemize}
\item word ngrams: presence or absence of contiguous sequences of 1, 2, 3, and 4 tokens; non-contiguous ngrams (ngrams with one token replaced by *);
\vspace*{-2mm}
\item character ngrams: presence or absence of contiguous sequences of 3, 4, and 5 characters;
\vspace*{-2mm}
%
\item all-caps: the number of words with all characters in upper case;
\vspace*{-2mm}
\item POS: the number of occurrences of each part-of-speech tag;
\vspace*{-2mm}
\item hashtags: the number of hashtags;
\vspace*{-2mm}
\item lexicons: the following sets of features were generated for each of the three manually constructed sentiment lexicons (NRC Emotion Lexicon, MPQA, Bing Liu Lexicon) and 
for each of the two automatically constructed lexicons (Hashtag Sentiment Lexicon and Sentiment140 Lexicon). 
Separate feature sets were produced for unigrams, bigrams, and non-contiguous pairs. 
The lexicon features were created for all tokens in the tweet, for each part-of-speech tag, for hashtags, and for all-caps tokens.	
For each token $w$ and emotion or polarity $p$, we used the sentiment/emotion score $score(w, p)$ to determine:
        \begin{itemize}
	         \item  total count of tokens in the tweet with $score(w, p) > 0$; 
	         \item  total score = $\sum_{w \in tweet} score(w, p)$;
	         \item  the maximal score = $max_{w \in tweet} score(w, p)$; 
	         \item  the score of the last token in the tweet with $score(w, p) > 0$; 
        \end{itemize}
\vspace*{-2mm}
\item punctuation: 
\begin{itemize}
	\item the number of contiguous sequences of exclamation marks, question marks, and both exclamation and question marks;
	\item whether the last token contains an exclamation or question mark;
\end{itemize}
\vspace*{-2mm}
\item emoticons: The polarity of an emoticon was determined with a regular expression
adopted from Christopher Potts' tokenizing script:\footnote{http://sentiment.christopherpotts.net/tokenizing.html}
\begin{itemize}
	\item presence or absence of positive and negative emoticons at any position in the tweet;
	\item whether the last token is a positive or negative emoticon;
\end{itemize}
\vspace*{-2mm}
\item elongated words: the number of words with one character repeated more than two times, for example,  `soooo';
\vspace*{-2mm}
\item clusters: The CMU pos-tagging tool provides the token clusters produced with the Brown clustering algorithm on 56 million English-language tweets. 
These 1,000 clusters serve as alternative representation of tweet content, reducing the sparcity of the token space. 
\begin{itemize}
	\item the presence or absence of tokens from each of the 1000 clusters;
\end{itemize}
\vspace*{-2mm}
\item negation: the number of negated contexts. Following \cite{PangLV02}, we defined a negated context as a segment of a tweet that starts with a negation word 
(e.g., {\it no, shouldn't}) and ends with one of the  punctuation marks: `,', `.', `:', `;', `!', `?'. 
A negated context affects the ngram and lexicon features: we add `\_NEG' suffix to each word following the negation word (`perfect' becomes `perfect\_NEG'). 
The `\_NEG' suffix is also added to polarity and emotion features (`POLARITY\_positive' becomes `POLARITY\_positive\_NEG').
The list of negation words was adopted from Christopher Potts' sentiment tutorial.\footnote{http://sentiment.christopherpotts.net/lingstruc.html}

\end{itemize}

\subsection{Experiments}
We trained the SVM classifier on the set of 9,912 annotated tweets (8,258 in the training set and 1,654 in the development set).
We applied the model to the test set of 3,813 unseen tweets. 
The same model was applied unchanged to the other test set of 2,094 SMS messages as well. 
The bottom-line score  used by the task organizers was the macro-averaged F-score of the positive and negative classes.
The results obtained by our system on the training set (ten-fold cross-validation), development set (when trained on the training set), and test sets 
(when trained on the combined set of tweets in the training and development sets)
are shown in Table~\ref{tab:TweetResults}.
The table also shows baseline results obtained by a majority classifier that always predicts the most frequent class as output.
Since the bottom-line F-score is based only on the F-scores of positive and negative classes (and not on neutral),
the majority baseline chose the most frequent class among positive and negative, which in this case was the positive class.
We also show baseline results obtained using an SVM and unigram features alone.
Our system (SVM and all features) obtained a macro-averaged F-score of 69.02 on the tweet set and 68.46 on the SMS set.
In the SemEval-2013 competition, our submission ranked first on both datasets. There were 48 submissions from 34 teams for this task.

%


Table \ref{tab:ablation-results} shows the results of the ablation experiments where we repeat the same classification process but remove one feature group at a time.
The most influential features for both datasets turned out to be the sentiment lexicon features: they provided gains of more than 8.5\%.
It is interesting to note that tweets benefited mostly from the automatic sentiment lexicons (NRC Hashtag Lexicon and the Sentiment140 Lexicon) whereas the SMS set benefited more from the manual lexicons (MPQA, NRC Emotion Lexicon, Bing Liu Lexicon).
Among the automatic lexicons, both the Hashtag Sentiment Lexicon and the Sentiment140 Lexicon contributed to roughly the same amount of improvement in performance on the tweet set.

The second most important feature group for the message-level task was that of ngrams (word and character ngrams). 
Expectedly, the impact of ngrams on the SMS dataset was less extensive since the classifier model was trained only on tweets.

 Attention to negations improved performance on both datasets.
 Removing the sentiment encoding features like hashtags, emoticons, and elongated words, had almost no impact on performance,
  but this is probably because the discriminating information in them was also captured by some other features such as character and word ngrams.

\begin{table}[t]
\caption{Message-level Task: The macro-averaged F-scores on different datasets.}
\label{tab:TweetResults}
\vspace*{-4mm}
\begin{center}
\resizebox{0.49\textwidth}{!}{
\begin{tabular}{llll}
\hline
 &{\bf Classifier} & {\bf Tweets} & {\bf SMS}\\
\hline
{\bf Training set:} 		&Majority	&26.94			&- \\
			 		&SVM-all 		& 67.20 & -\\ [4pt]
{\bf Development set:} 	&Majority 	& 26.85 &-\\
					&SVM-all		& 68.72 & -\\ [4pt]
{\bf Test set:}		&Majority	& 29.19 &19.03\\
					&SVM-unigrams	& 39.61 &39.29\\
					&SVM-all  		& {\bf 69.02} & {\bf 68.46}\\
\hline
\end{tabular}
}
\end{center}
\vspace*{-6mm}
\end{table}

\begin{table}[t!]
\caption{\label{tab:ablation-results} Message-level Task: The macro-averaged F-scores obtained on the test sets with one of the feature groups removed.
 The number in the brackets is the difference with the {\it all features} score.
The biggest drops are shown in bold.}
\vspace*{-4mm}
\begin{center}
\resizebox{0.49\textwidth}{!}{
\begin{tabular}{llll}
\hline \bf Experiment & \bf Tweets & & \bf SMS \\ \hline
all features &  69.02 & &   68.46 \\[5pt]
all - lexicons &    60.42 ({\bf -8.60}) &  & 59.73 ({\bf -8.73})\\
$\,\,\,\,$ all - manual lex.    & 67.45 (-1.57) & & 65.64 (-2.82)\\
$\,\,\,\,$ all - auto.\@ lex.   & 63.78 (-5.24) & & 67.12 (-1.34)\\
$\,\,\,\,$ all - Senti140 lex.\@    & 65.25 (-3.77) & & 67.33 (-1.13)\\
$\,\,\,\,$ all - Hashtag lex.   & 65.22 (-3.80) & & 70.28 (1.82)\\ [5 pt]

all - ngrams & 61.77 (-7.25) & & 67.27 (-1.19) \\
$\,\,\,\,$ all - word ngrams   & 64.64 (-4.38) & & 66.56 (-1.9)\\
$\,\,\,\,$ all - char.\@ ngrams    & 67.10 (-1.92) & & 68.94 (0.48) \\ [5 pt]
all - negation  & 67.20 (-1.82) & & 66.22 (-2.24)\\
all - POS   & 68.38 (-0.64) & & 67.07 (-1.39)\\
all - clusters	& 69.01 (-0.01)	& & 68.10 (-0.36)\\
\multicolumn{4}{l}{all - encodings (elongated, emoticons, punctuations,}\\
all-caps, hashtags) & 69.16 (0.14) & & 68.28 (-0.18)\\

\hline
\end{tabular}
}
\end{center}
\vspace*{-6mm}
\end{table}

\section{Task: Automatically Detecting the Sentiment of a Term in a Message}
The objective of this task is to detect whether a term (a word or phrase) within a message conveys a positive, negative, or neutral sentiment.
Note that the same term may express different sentiments in different contexts.

\subsection{Classifier and features}
We trained an SVM using the LibSVM package \cite{Chang2011} and a linear kernel.
In ten-fold cross-validation over the training data, the linear kernel 
outperformed other kernels implemented in LibSVM as well as a maximum-entropy classifier.
Our model leverages a variety of features, as described below:



\vspace*{-2mm}
\begin{itemize}
\item word ngrams: 
\vspace*{-1mm}
\begin{itemize}
\item presence or absence of unigrams, bigrams, and the full word string of a target term;
\item leading and ending unigrams and bigrams;
\end{itemize}
\vspace*{-1mm}
	\item character ngrams: presence or absence of two- and three-character prefixes and suffixes of all the words in a target term 
(note that the target term may be a multi-word sequence);
\vspace*{-1mm}
	\item elongated words: presence or absence of elongated words (e.g., 'sooo');
\vspace*{-1mm}
	\item emoticons: the numbers and categories of emoticons that a term contains\footnote{http://en.wikipedia.org/wiki/List\_of\_emoticons};
\vspace*{-1mm}
	\item punctuation: presence or absence of punctuation sequences such as `?!' and `!!!';
\vspace*{-1mm}
	\item upper case: 
\vspace*{-1mm}
	\begin{itemize}
      \item whether all the words in the target start with an upper case letter followed by lower case letters;
      \item whether the target words are all in uppercase (to capture a potential named entity);
	\end{itemize}
\vspace*{-1mm}
	\item stopwords: whether a term contains only stop-words. If so, separate features indicate whether there are 1, 2, 3, or more stop-words;
\vspace*{-1mm}
	\item lengths: 
\begin{itemize}
\item the length of a target term (number of words); 
\item the average length of words (number of characters) in a term; 
\item a binary feature indicating whether a term contains long words;
\end{itemize}
\vspace*{-1mm}
	\item negation: similar to those described for the message-level task.  
	Whenever a negation word was found immediately before the target or within the target, the polarities of all tokens after the negation term were flipped;
\vspace*{-1mm}
	\item position: whether a term is at the beginning, end, or another position;
\vspace*{-1mm}
	\item sentiment lexicons: we used automatically created lexicons (NRC Hashtag Sentiment Lexicon, Sentiment140 Lexicon) as well as manually created lexicons (NRC Emotion Lexicon, MPQA, Bing Liu Lexicon). 
\vspace*{-1mm}
	\begin{itemize}
      \item total count of tokens in the target term with sentiment score greater than 0;
      \item the sum of the sentiment scores for all tokens in the target;
      \item the maximal sentiment score;
      \item the non-zero sentiment score of the last token in the target; 
	\end{itemize}
\vspace*{-2mm}
	\item term splitting: when a term contains a hashtag made of multiple words (e.g., \#biggestdaythisyear), we split the hashtag into component words;
\vspace*{-1mm}
\item others: 
\vspace*{-1mm}
	\begin{itemize}
      \item whether a term contains a Twitter user name;
      \item whether a term contains a URL.
	\end{itemize}
\end{itemize}

The above features were extracted from target terms as well as from the rest of the message (the context).
For unigrams and bigrams, we used four words on either side of the target as the context. The window size was chosen through experiments on the development set.

\subsection{Experiments}

We trained an SVM classifier on the 8,891 annotated terms in tweets (7,756 terms in the training set and 1,135 terms in the development set).
We applied the model to 4,435 terms in the tweets test set.
The same model was applied unchanged to the other test set of 2,334 terms in unseen SMS messages as well.
The bottom-line score  used by the task organizers was the macro-averaged F-score of the positive and negative classes.

The results on the training set (ten-fold cross-validation), the development set (trained on the training set), and the test sets
(trained on the combined set of tweets in the training and development sets)
are shown in Table~\ref{tab:TermResults}.
The table also shows baseline results obtained by a majority classifier that always predicts the most frequent class as output,
and an additional baseline result obtained using an SVM and unigram features alone.
Our submission obtained a macro-averaged F-score of 88.93 on the tweet set and was ranked first among 29 submissions from 23 participating teams.
Even with no tuning specific to SMS data, our SMS submission still obtained second rank with an F-score of 88.00.
The score of the first ranking system on the SMS set was 88.39.
A post-competition bug-fix in the bigram features resulted in a small improvement: F-score of 89.10 on the tweets set and 88.34 on the SMS set.

Note that the performance is significantly higher in the term-level task than in the message-level task.
 This is largely because of the ngram features (see unigram baselines
 in Tables 2 and 4).
We analyzed the labeled data provided to determine why ngrams performed so strongly in this task.
We found that the percentage of test tokens already
 seen within training data targets was 85.1\%. 
Further, the average ratio of instances pertaining to the most dominant polarity of a target term to the total number of instances
 of that target term was 0.808.

Table \ref{tab:TermAblation} presents the ablation F-scores.
Observe that the ngram features were the most useful. 
Note also that removing just the word ngram features or just the character ngram features results in only a small drop
in performance. This indicates that the two feature groups capture similar information. 

The sentiment lexicon features are the next most useful group---removing them leads to a drop in F-score of
3.95 points for the tweets set and 4.64 for the SMS set.
Modeling negation improves the F-score by 0.72 points on the tweets set and 1.57 points on the SMS set. 

The last two rows in Table \ref{tab:TermAblation} show the results obtained when the features are extracted only from the target (and not from its context)
and when they are extracted only from the context of the target (and not from the target itself).
Observe that even though the context may influence the polarity of the target, using target features alone
is substantially more useful than using context features alone.
Nonetheless, adding context features improves the F-scores by roughly 2 to 4 points. 

\begin{table}[t]
\caption{Term-level Task: The macro-averaged F-scores on the datasets. The official scores of our submission are shown in bold. SVM-all* shows results after a bug fix.}
\label{tab:TermResults}
\begin{center}
\vspace*{-4mm}
\resizebox{0.49\textwidth}{!}{
\begin{tabular}{llll}
\hline
 &{\bf Classifier} & {\bf Tweets} & {\bf SMS}\\
\hline
{\bf Training set:} 		&Majority	& 38.38			&- \\
			 		&SVM-all 		& 86.80 & -\\ [4pt]
{\bf Development set:} 	&Majority 	& 36.34 &-\\
					&SVM-all		& 86.49 & -\\ [4pt]
{\bf Test set:}		&Majority	& 38.13 &32.11\\
					&SVM-unigrams	&  80.28                &78.71 \\
					&official SVM-all  		& {\bf 88.93} & {\bf 88.00}\\
					&SVM-all*  		& 89.10 & 88.34\\
\hline
\end{tabular}
}
\end{center} 
\vspace*{-4mm}
\end{table}

\begin{table}[t!]
\caption{\label{tab:TermAblation} Term-level Task: The F-scores obtained on the test sets with one of the feature groups removed.
The number in brackets is the difference with the {\it all features} score.
The biggest drops are shown in bold.}
\begin{center}
\vspace*{-4mm}
\resizebox{0.49\textwidth}{!}{
\begin{tabular}{llll}
\hline \bf Experiment 		&\bf Tweets 	& & \bf SMS \\ \hline
all features                &89.10          & & 88.34\\[4pt]
all - ngrams                &83.86 ({\bf -5.24})    & &80.49 ({\bf -7.85})\\
\;\;\; all - word ngrams    &88.38 (-0.72)  & &87.37 (-0.97)\\
\;\;\; all - char.\@ ngrams &89.01 (-0.09)   & &87.31 (-1.03)\\[4pt]

all - lexicons              &85.15 (-3.95)  & &83.70 (-4.64)\\
\;\;\;all - manual lex.     &87.69 (-1.41)  & &86.84 (-1.5)\\
\;\;\;all - auto lex.       &88.24 (-0.86)  & &86.65 (-1.69)\\[4pt]

all - negation              &88.38 (-0.72)  & &86.77 (-1.57)\\
all - stopwords             &89.17 (0.07)   & &88.30 (-0.04)\\
\multicolumn{4}{l}{all - encodings (elongated words, emoticons, punctns.,} \\
\;\;\;\;\;\; uppercase)     &89.16 (0.06)   & &88.39 (0.05)\\[4pt]
\hline
& & & \\ [-9pt]
all - target                &72.97 ({\bf -16.13}) & &68.96 ({\bf -19.38})\\
all - context               &85.02 (-4.08)  & &85.93 (-2.41)\\


\hline
\end{tabular}
}
\end{center}
\vspace*{-5mm}
\end{table}

\section{Conclusions}

We created two state-of-the-art SVM classifiers,
one to detect the sentiment of messages and one to detect the
sentiment of a term within a message. 
Our submissions on tweet data stood first in both these subtasks of the SemEval-2013 competition `Detecting Sentiment in Twitter'.
We implemented a variety of features based on surface form and lexical categories.
The sentiment lexicon features (both manually created and automatically generated) along with ngram features (both word and character ngrams) 
led to the most gain in performance. 

\section*{Acknowledgments}
We thank Colin Cherry for providing his SVM code and for helpful discussions.

\bibliography{references}

\begin{thebibliography}{}

\bibitem[\protect\citename{Chang and Lin}2011]{Chang2011}
Chih-Chung Chang and Chih-Jen Lin.
\newblock 2011.
\newblock {LIBSVM}: A library for support vector machines.
\newblock {\em ACM Transactions on Intelligent Systems and Technology},
  2(3):27:1--27:27.

\bibitem[\protect\citename{Chew and Eysenbach}2010]{Chew10}
Cynthia Chew and Gunther Eysenbach.
\newblock 2010.
\newblock {Pandemics in the Age of Twitter: Content Analysis of Tweets during
  the 2009 H1N1 Outbreak}.
\newblock {\em PLoS ONE}, 5(11):e14118+, November.

\bibitem[\protect\citename{Fan \bgroup et al.\egroup }2008]{liblinear}
R.-E. Fan, K.-W. Chang, C.-J. Hsieh, X.-R. Wang, and Lin C.-J.
\newblock 2008.
\newblock {LIBLINEAR: A Library for Large Linear Classification}.
\newblock {\em Journal of Machine Learning Research}, 9:1871--1874.

\bibitem[\protect\citename{Gimpel \bgroup et al.\egroup }2011]{Gimpel11}
Kevin Gimpel, Nathan Schneider, Brendan O'Connor, Dipanjan Das, Daniel Mills,
  Jacob Eisenstein, Michael Heilman, Dani Yogatama, Jeffrey Flanigan, and
  Noah~A. Smith.
\newblock 2011.
\newblock {Part-of-Speech Tagging for Twitter: Annotation, Features, and
  Experiments}.
\newblock In {\em Proceedings of the Annual Meeting of the Association for
  Computational Linguistics}.

\bibitem[\protect\citename{Go \bgroup et al.\egroup }2009]{Go2009}
Alec Go, Richa Bhayani, and Lei Huang.
\newblock 2009.
\newblock {Twitter Sentiment Classification using Distant Supervision}.
\newblock In {\em Final Projects from CS224N for Spring 2008/2009 at The
  Stanford Natural Language Processing Group}.

\bibitem[\protect\citename{Hu and Liu}2004]{Hu04}
Minqing Hu and Bing Liu.
\newblock 2004.
\newblock Mining and summarizing customer reviews.
\newblock In {\em Proceedings of the tenth ACM SIGKDD international conference
  on Knowledge discovery and data mining}, KDD '04, pages 168--177, New York,
  NY, USA. ACM.

\bibitem[\protect\citename{Jansen \bgroup et al.\egroup }2009]{Jansen09}
Bernard~J. Jansen, Mimi Zhang, Kate Sobel, and Abdur Chowdury.
\newblock 2009.
\newblock Twitter power: Tweets as electronic word of mouth.
\newblock {\em Journal of the American Society for Information Science and
  Technology}, 60(11):2169--2188.

\bibitem[\protect\citename{Mandel \bgroup et al.\egroup }2012]{Mandel12}
Benjamin Mandel, Aron Culotta, John Boulahanis, Danielle Stark, Bonnie Lewis,
  and Jeremy Rodrigue.
\newblock 2012.
\newblock A demographic analysis of online sentiment during hurricane irene.
\newblock In {\em Proceedings of the Second Workshop on Language in Social
  Media}, LSM '12, pages 27--36, Stroudsburg, PA, USA. Association for
  Computational Linguistics.

\bibitem[\protect\citename{Mohammad and Turney}2010]{MohammadT10}
Saif~M. Mohammad and Peter~D. Turney.
\newblock 2010.
\newblock {Emotions Evoked by Common Words and Phrases: Using Mechanical Turk
  to Create an Emotion Lexicon}.
\newblock In {\em Proceedings of the NAACL-HLT 2010 Workshop on Computational
  Approaches to Analysis and Generation of Emotion in Text}, LA, California.

\bibitem[\protect\citename{Mohammad and Yang}2011]{MohammadY11}
Saif Mohammad and Tony Yang.
\newblock 2011.
\newblock {Tracking Sentiment in Mail: How Genders Differ on Emotional Axes}.
\newblock In {\em Proceedings of the 2nd Workshop on Computational Approaches
  to Subjectivity and Sentiment Analysis (WASSA 2.011)}, pages 70--79,
  Portland, Oregon. Association for Computational Linguistics.

\bibitem[\protect\citename{Mohammad}2012]{Mohammad12}
Saif Mohammad.
\newblock 2012.
\newblock {\#Emotional Tweets}.
\newblock In {\em Proceedings of the First Joint Conference on Lexical and
  Computational Semantics (*SEM)}, pages 246--255, Montr\'eal, Canada.
  Association for Computational Linguistics.

\bibitem[\protect\citename{Pang \bgroup et al.\egroup }2002]{PangLV02}
Bo~Pang, Lillian Lee, and Shivakumar Vaithyanathan.
\newblock 2002.
\newblock {Thumbs up?: Sentiment Classification Using Machine Learning
  Techniques}.
\newblock In {\em Proceedings of the Conference on Empirical Methods in Natural
  Language Processing}, pages 79--86, Philadelphia, PA.

\bibitem[\protect\citename{Salath{\'e} and Khandelwal}2011]{Salathe11}
Marcel Salath{\'e} and Shashank Khandelwal.
\newblock 2011.
\newblock Assessing vaccination sentiments with online social media:
  Implications for infectious disease dynamics and control.
\newblock {\em PLoS Computational Biology}, 7(10).

\bibitem[\protect\citename{Verma \bgroup et al.\egroup }2011]{Verma11}
Sudha Verma, Sarah Vieweg, William Corvey, Leysia Palen, James Martin, Martha
  Palmer, Aaron Schram, and Kenneth Anderson.
\newblock 2011.
\newblock Natural language processing to the rescue? extracting "situational
  awareness" tweets during mass emergency.
\newblock In {\em International AAAI Conference on Weblogs and Social Media}.

\bibitem[\protect\citename{Wilson \bgroup et al.\egroup }2005]{Wilson05}
Theresa Wilson, Janyce Wiebe, and Paul Hoffmann.
\newblock 2005.
\newblock Recognizing contextual polarity in phrase-level sentiment analysis.
\newblock In {\em Proceedings of the conference on Human Language Technology
  and Empirical Methods in Natural Language Processing}, HLT '05, pages
  347--354, Stroudsburg, PA, USA. Association for Computational Linguistics.

\bibitem[\protect\citename{Wilson \bgroup et al.\egroup
  }2013]{SemEval2013task2}
Theresa Wilson, Zornitsa Kozareva, Preslav Nakov, Sara Rosenthal, Veselin
  Stoyanov, and Alan Ritter.
\newblock 2013.
\newblock {SemEval}-2013 task 2: Sentiment analysis in twitter.
\newblock In {\em Proceedings of the International Workshop on Semantic
  Evaluation}, SemEval '13, Atlanta, Georgia, USA, June.

\end{thebibliography}

\end{document}